\documentclass[conference]{IEEEtran}
\IEEEoverridecommandlockouts
\usepackage{cite}
\usepackage{amsmath,amssymb,amsfonts}
\usepackage{algorithmic}
\usepackage{graphicx}
\usepackage{textcomp}
\usepackage[table]{xcolor}
\usepackage{xcolor}
\usepackage{booktabs}
\usepackage{makecell}
\usepackage{etoolbox}

\makeatletter

\patchcmd{\@makecaption}
  {\scshape}
  {}
  {}
  {}
\makeatother
\usepackage{soul}
\usepackage{diagbox}
\usepackage[font=footnotesize]{caption}

\def\BibTeX{{\rm B\kern-.05em{\sc i\kern-.025em b}\kern-.08em
    T\kern-.1667em\lower.7ex\hbox{E}\kern-.125emX}}
\begin{document}

\title{Spike-NeRF: Neural Radiance Field Based On Spike Camera}
\author{Yijia Guo\textsuperscript{\rm 1}, 
Yuanxi Bai\textsuperscript{\rm 2},
Liwen Hu\textsuperscript{\rm 1},
Mianzhi Liu\textsuperscript{\rm 2},
Ziyi Guo\textsuperscript{\rm 2},
Lei Ma\textsuperscript{\rm 1,2*}\thanks{* Corresponding author.

This paper is accepted by ICME2024},
Tiejun Huang\textsuperscript{\rm 1}\\
\textsuperscript{\rm 1}National Engineering Research Center of Visual Technology (NERCVT), Peking University\\
\textsuperscript{\rm 2}College of Future Technology, Peking University\\
}


\maketitle
\begin{abstract}
As a neuromorphic sensor with high temporal resolution, spike cameras offer notable advantages over traditional cameras in high-speed vision applications such as high-speed optical estimation, depth estimation, and object tracking. Inspired by the success of the spike camera, we proposed Spike-NeRF, the first Neural Radiance Field derived from spike data, to achieve 3D reconstruction and novel viewpoint synthesis of high-speed scenes. Instead of the multi-view images at the same time of NeRF, the inputs of Spike-NeRF are continuous spike streams captured by a moving spike camera in a very short time. To reconstruct a correct and stable 3D scene from high-frequency but unstable spike data, we devised spike masks along with a distinctive loss function. We evaluate our method qualitatively and numerically on several challenging synthetic scenes generated by blender with the spike camera simulator. Our results demonstrate that Spike-NeRF produces more visually appealing results than the existing methods and the baseline we proposed in high-speed scenes. Our code and data will be released soon.
\end{abstract}
%
%
    
\section{Introduction}
\label{sec:intro}
Novel-view synthesis (NVS) is a long-standing problem aiming to render photo-realistic images from novel views of a scene from a sparse set of input images.
This topic has recently seen impressive progress due to the use of neural networks to learn
representations that are well suited for view synthesis tasks, known as Neural Radiance Field (NeRF)\cite{nerf,barron2021mipnerf,barron2022mipnerf360,martin2021nerfwild,chen2022tensorf,wang2021nerf--,zhang2020nerf++}. 
Despite its success, NeRF performs awfully in high-speed scenes since the motion blur caused by high-speed scenes violates the assumption by NeRF that the input images are sharp. Deblur methods such as Deblur-NeRF \cite{ma2022deblur} and BAD-NeRF\cite{wang2023bad} can only handle mild motion blur. The introduction of high-speed neuromorphic cameras, such as event cameras \cite{liu2022devrf} and spike cameras, are expected to fundamentally solve this problem. 

Spike camera \cite{joshi2017retina,dong2021spike} is a neuromorphic sensor, where each pixel captures photons independently, keeps recording the luminance intensity asynchronously, and outputs binary spike
streams to record the dynamic scenes at extremely high temporal resolution (40000Hz).  Recently, many existing approaches use spike data to reconstruct image \cite{zhu2021neuspike,zhao2021spk2imgnet,zhang2023learning} for high-speed scenes, or directly perform downstream tasks such as optical flow estimation \cite{hu2022optical}
 and depth estimation \cite{wang2022learning}.
 
 Motivated by the notable success achieved by spike cameras and NeRF's
 \begin{figure}[htbp]
    \centering
    \includegraphics[width=\linewidth]{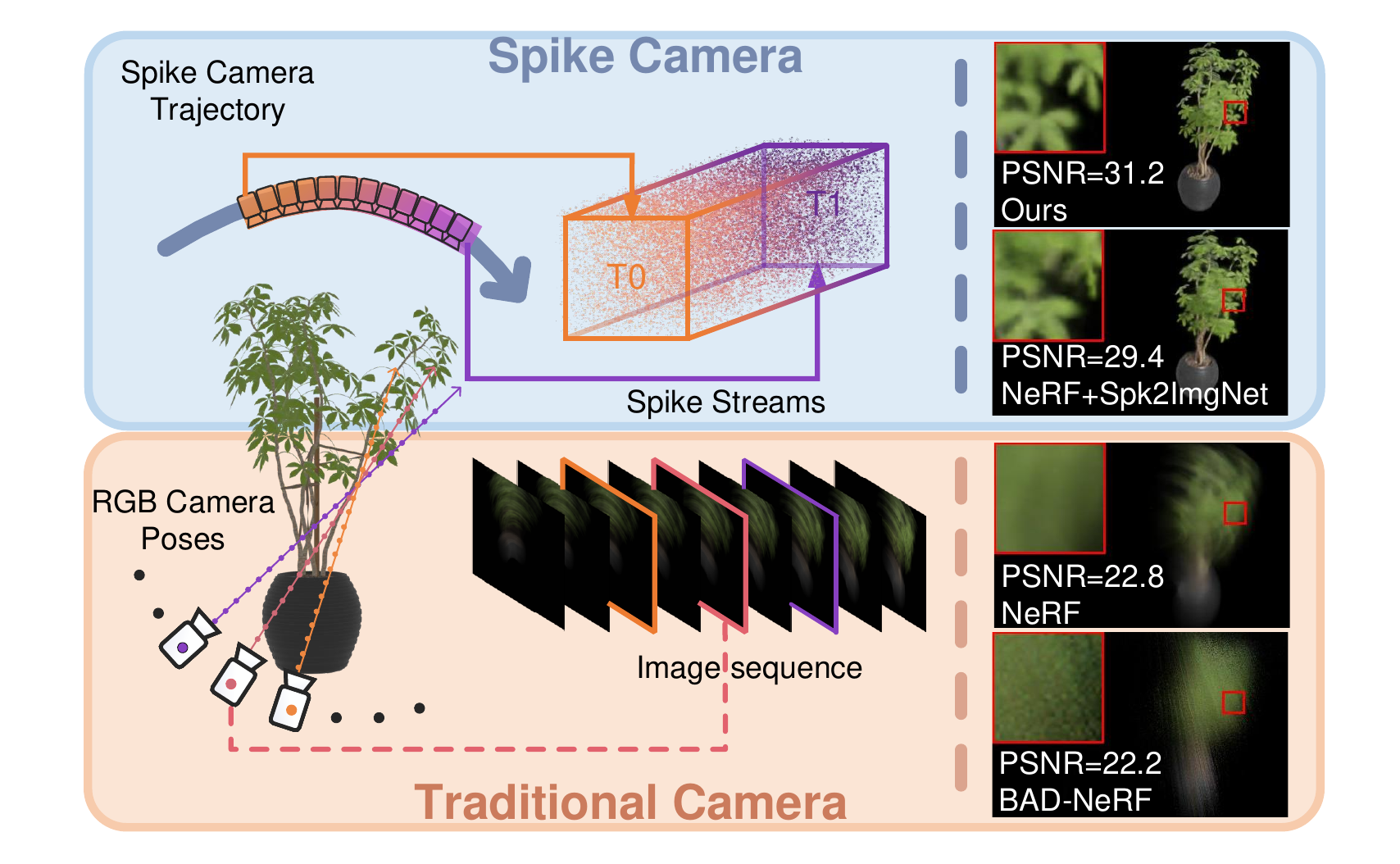}
    \caption{ Existing works on NeRF (orange background) are reconstructed from image sequences generated by traditional cameras which record the luminance intensity during the exposure time at a fixed frame rate producing strong blur in high-speed scenes. Our approach (blue background) produces significantly better and sharper results by using dense spike streams instead of image sequences.}\label{f1_2}
\end{figure}
 limitation, we proposed Spike-NeRF, the first Neural Radiance Field built by spike data. Different from NeRF, we use a set of continuously spike streams as inputs instead of images from different perspectives at the same time (see Figure \ref{f1_2}). To reconstruct a volumetric 3D representation of a 
 scene from spike streams and generate a new spike stream for novel views based on this scene, we first proposed a spiking volume renderer based on the coding method of spike cameras. It generates spike streams asynchronously with radiance obtained by ray casting. Additionally, we both use spike loss to reduce local blur and spike masks to limit NeRF learning information in a specific area, thereby mitigating artifacts resulting from reconstruction errors and noise. 

Our experimental results show that Spike-NeRF is suitable for high-speed scenes that would not be conceivable with traditional cameras. Moreover, our method is largely superior to directly using reconstructed images of spike streams for supervision which is considered as the baseline. Our main contributions can be summarized as follows:

\begin{itemize}
\item[$\bullet$] Spike-NeRF, the first approach for inferring NeRF from
a spike stream that enables a novel view synthesis in both gray and RGB space for high-speed scene.

\item[$\bullet$]  A bespoke rendering strategy for spike streams leading to data-efficient training and spike stream generating.

\item[$\bullet$] A dataset containing RGB spike data and high-frequency (40,000fps) camera poses
\end{itemize}

\section{Related Work}
\label{sec:related_work}
\subsection{NeRF on traditional cameras}
Neural Radiance Field (NeRF) \cite{nerf} has arisen as a significant development in the field of Computer Vision and Computer Graphics, used for synthesizing novel views of a scene from a sparse set of images by combining machine learning with geometric reasoning. Various research and approaches based on NeRF have been proposed recently. For example, \cite{park2021nerfies,pumarola2021dnerf,li2021SceneFlownerf,yan2023nerfds} extend NeRF to dynamic and non-grid scenes, \cite{barron2021mipnerf,barron2022mipnerf360} significantly improve the rendering quality of NeRF and \cite{wang2023bad,ma2022deblur,lee2023dp,lee2023exblurf}  robustly severe blurred images that could affect the rendering quality of NeRF.

\begin{figure*}[htbp]
\includegraphics[width=2.0\columnwidth]{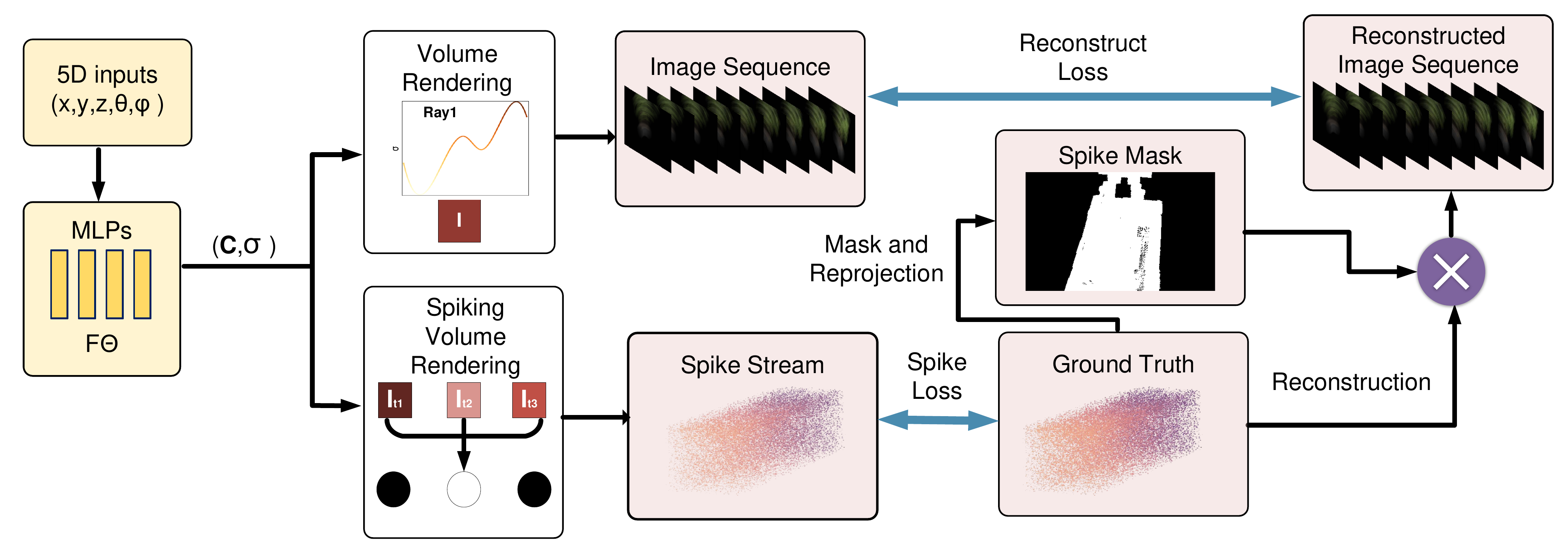}
\centering
\caption{Overview of our Spike-NeRF. Same as NeRF, we use color(\textbf{C}) and density(\textbf{$\sigma$}) generated by MLPs as the input of volume renderer (equation  (\ref{nerf:qiuhe}))  and spiking volume renderer (equation (\ref{spikenerf:qiuhe})). We proposed reconstruct loss between the volume renderer result and masked images which are reconstructed from g.t. spike streams. Spike loss between the spike rendering result generated by our spiking volume renderer and g.t. spike streams is computed too.}\label{overview}
\end{figure*}

\subsection{NeRF on Neuromorphic Cameras}
Neuromorphic sensors have shown their advantages in most computer version problems including novel views synthesizing. EventNeRF \cite{rudnev2023eventnerf} and Ev-nerf \cite{hwang2023ev} synthesize the novel view in scenarios such as high-speed movements that would not be conceivable with a traditional camera by event supervision. Nonetheless, these works assume that event streams are temporally dense and low-noise which is inaccessible in practice. Robust e-NeRF \cite{low2023robust} incorporates a more realistic event generation model to directly and robustly reconstruct NeRFs under various real-world conditions. DE-NeRF \cite{ma2023deformable} and E2NeRF \cite{qi2023e2nerf} extend event nerf to dynamic scenes and severely blurred images as NeRF researchers did.

\subsection{Spike Camera Application}
As a neuromorphic sensor with high temporal resolution, spike cameras \cite{dong2021spike}  offer significant advantages on many high-speed version tasks. \cite{zheng2021high} and \cite{zhu2021neuspike} propose spike stream reconstruction methods for high-speed scenes. Subsequently, deep learning-based reconstruction frameworks \cite{zhao2021spk2imgnet,zhang2023learning} are introduced to reconstruct spike streams robustly. Spike cameras also show its superiors on downstream tasks such as optical flow estimation \cite{hu2022optical,chen2023self}, monocular and stereo depth estimation \cite{wang2022learning,zhang2022spike}, Super-resolution \cite{zhao2023learning} and high-speed real-time object tracking \cite{zheng2022spike}.
\section{Preliminary}
\label{sec:related_work}
\subsection{Spike Camera And Its Coding Method}
Unlike traditional cameras that record the luminance intensity of each pixel during the exposure time at a fixed frame rate, tensors on spike cameras of each pixel capture photons independently and keep recording the luminance intensity asynchronously without a dead zone. 

Each pixel on the spike camera converts the light signal into a current signal. When the accumulated intensity reaches the dispatch threshold, a spike is fired and the accumulated intensity is reset. For pixel  $\boldsymbol{x} = (x, y)$,this process can be expressed as
\begin{align}
    {A}(\boldsymbol{x}, t) &= {A}(\boldsymbol{x}, t-1) + {I}(\boldsymbol{x}, t) \label{eq:ax} 
\end{align}
\begin{align}
    {s}(\boldsymbol{x}, t) = 
        \begin{cases}
        1 & \text{if } {A}(\boldsymbol{x}, t-1) + {I}(\boldsymbol{x}, t) > \phi \\
        0 & \text{otherwise}
        \end{cases} \label{eq:sx}
\end{align}

where:
   \begin{align}
    {I}(\boldsymbol{x}, t)= \int_{t-1}^{t} {I_{in}}(\boldsymbol{x}, \tau) d\tau\ {\rm{mod}} \; \phi
    \end{align}\label{sx1}
Here ${A}(\boldsymbol{x}, t)$ is the accumulated intensity at time $t$, ${s}(\boldsymbol{x}, t)$ is the spike output at time $t$ and ${I_{in}}(\boldsymbol{x}, \tau)$ is the input current at time $\tau$ (proportional to light intensity). We will directly use $I(\boldsymbol{x}, t)$ to represent the luminance intensity to simplify our presentation. Further, due to the limitations of circuits, each spike is read out at discrete time $nT, n \in \mathbb{N}$ ($T$ is a micro-second level). Thus, the output of the spike camera is a spatial-temporal binary stream $S$ with $H \times W \times N$ size. Here, $H$ and $W$ are the height and width of the sensor, respectively, and $N$ is the temporal window size of the spike stream.

\subsection{Neural Radiance Field (NeRF) Theory}

Neural Radiance Field (NeRF) uses a 5D vector-valued function to represent a continuous scene. The input to this function consists of a 3D location \textbf{x} = (x, y, z) and 2D viewing direction \textbf{d} =($\theta$,$\phi$),
while output is an emitted color \textbf{c} = (r, g, b) and volume density  $\sigma$.  Both $\sigma$ and
\textbf{c} are represented implicitly as multi-layer perceptrons (MLPs), written as:
\begin{equation}
		F_\Theta:(\textbf{x},\textbf{d})\to(\textbf{c},\sigma)\label{nerf:F}
\end{equation}
Given the volume density $\sigma$ and color functions \textbf{c}, the rendering result $I$ of any given ray $\textbf{r}=\textbf{o}+t\textbf{d}$ passes through the scene can be computed using principles from volume rendering.
\begin{equation}
		I(\textbf{r})=\begin{matrix} \int_{t_n}^{t_f} T(t)\sigma(\textbf{r}(t))\textbf{c}(\textbf{r}(t),\textbf{d})\, dt\end{matrix}\label{nerf:jifeng}
\end{equation}
where
\begin{equation}
	T(t)=e^{-\begin{matrix} \int_{t_n}^{t} \sigma(\textbf{r}(s))\, ds\end{matrix}}\label{nerf:t}
\end{equation}
The function T(t) denotes the accumulated transmittance along the ray from $t_n$ to t. For computing reasons, rays were divided into N equally spaced bins, and a sample
was uniformly drawn from each bin. Then, equation \ref{nerf:jifeng} can
be approximated as
\begin{equation}
		I(\textbf{r})=\begin{matrix} \sum_{i=1}^{N} T_i(t)(1-\exp(-\sigma_i\delta_i))\textbf{c}_i\label{nerf:qiuhe}
\end{matrix}
\end{equation}
where:
\begin{equation}
		T_i(t)=\begin{matrix} \sum_{j=1}^{i-1} -\exp(-\sigma_i\delta_i)\textbf{c}_i\label{nerf:t_qiuhe}
\end{matrix}
\end{equation}
and:
\begin{equation}
		\delta_i=t_{i+1}-t_i
\end{equation}
After calculating the color $I(\textbf{r})$ for each pixel, a square error photometric loss is used to
optimize the MLP parameters.
\begin{equation}
		L= \sum_{r \in R} \Vert I(\textbf{r})-I_{gt}(\textbf{r}) \Vert\label{nerf:loss}
\end{equation}

\section{Method}
\label{sec:method}
\subsection{Overview}
Taking inspiration from NeRF, Spike-NeRF implicitly represents the static scenes as an MLP network $F_\Theta$ with 5D inputs:
\begin{equation}
		F_\Theta:(\textbf{x($t_i$)},\textbf{d($t_i$)})\to(\textbf{c},\sigma)\label{sp_nerf:F}
\end{equation}
Here, each $t_i$ corresponds to a frame of spike $s_i=\{0,1\}$ in the continuous spike stream $\mathbb{S} = \{ {s}_{i} \in \mathbb{R}^{W \times H} | i = 0,  1,  2, \dots\}$ captured by a spike camera in a very short time. Considering the difficulty of directly using spike streams for supervision, we firstly reconstruct the spike stream $\mathbb{S}$ into image sequence $\mathbb{I}= \{ {im}_{i} \in \mathbb{R}^{W \times H} | i = 0,  1,  2, \dots\}$ where $ {im}_{i}$ is the reconstructed image at $t_i$. We use the results with inputs=$\mathbb{I}$ as our baseline. Since all methods reconstruct images from multi-frame spikes, using the reconstructed images as a supervision signal would lead to artifacts and blurring. We introduce spike masks ${M_s}$ to make NeRF focus on the triggered area. We also propose a spiking volume renderer based on the coding method of the spike camera to generate spike streams for novel views. We then use the g.t. spike streams constraint network directly.
\par The total loss used to train Spike-NeRF is given by:
\begin{equation}
L_{total}=L_{recon}+{\lambda}L_{spike}
\end{equation}
$L_{recon}$ is the loss between the image rendering result and masked images which are reconstructed from g.t. spike streams. $L_{spike}$ is the loss between the spike rendering result generated by our spiking volume rendering method and g.t. spike streams.




\subsection{Spiking Volume Renderer}
If we introduce time $t$ into the volume rendering equation \ref{nerf:jifeng}, the rendering results $I(\textbf{r},t)$ of any given ray $\textbf{r(t)}=\textbf{o(t)}+k\textbf{d(t)}$ at time $t$ is:
\begin{equation}
		I(\textbf{r},t)=\begin{matrix} \int_{k_n}^{k_f} T(k,t)\sigma(\textbf{r}(k,t))\textbf{c}(\textbf{r}(k,t),\textbf{d}(t))\, dk\end{matrix}\label{snerf:jifeng}
\end{equation}
Where
\begin{equation}
	T(k,t)=e^{-\begin{matrix} \int_{k}^{k_n} \sigma(\textbf{r}(s,t))\, ds\end{matrix}}\label{snerf:t}
\end{equation}
Then, if we assume that for any $x$ $A(x,t_0)=0$, equation \ref{ax} can be written as:
\begin{equation}
		A(x,t)=\begin{matrix}\int_{t_0}^{t}I(x,t)dt\end{matrix}-N\phi\label{snerf:jifeng_t}
\end{equation}
Here,$\phi$ is the threshold of the spike camera and N is the number of "1" for spike streams $\mathbb{S}(x) = \{ {s}_{t_i}|t_i\in (t_0,t) \}$ and x=(x,y) is the coordinates for each pixel. 
For computing reasons, rays were divided into $N_0$ equally spaced bins, $(t_0,t)$ were divided into $N_1$ equally spaced bins, and a sample was uniformly drawn from each bin. Then, equation 2 can
 be written as:
\begin{align}
    {s}(\boldsymbol{x}, t) = 
    \begin{cases}
        1 & \text{if } \sum_{i=1}^{N_1} I(x,t_i) - N\phi > \phi \\
        0 & \text{otherwise}
    \end{cases} \label{eq:sx_nerf}
\end{align}

where:
\begin{equation}
		I(\textbf{r},t)=\begin{matrix} \sum_{i=1}^{N_0} T_i(k,t)(1-\exp(-\sigma_i\delta_i))\textbf{$c_i$}(t)\label{spikenerf:qiuhe}
\end{matrix}
\end{equation}
where:
\begin{equation}
		T_i(k,t)=\begin{matrix} \sum_{j=1}^{i-1} -\exp(-\sigma_i\delta_i)\textbf{$c_i$}(t)\label{spikenerf:t_qiuhe}
\end{matrix}
\end{equation}
and:
\begin{equation}
		\delta_i=t_{i+1}-t_i
\end{equation}
However, $A(x,t_0)$ is not equal to 0 in real situations. To address this, we introduce a random startup matrix and utilize the stable results after several frames. The above processes do not participate in backpropagation as they are not differentiable. After generating spike streams $\mathbb{S}$, we can compute:
\begin{equation}
		L_{spike}= \sum_{r \in R} \Vert\mathbb{S}(\textbf{x})-\mathbb{S}_{gt}(\textbf{x}) \Vert\label{nerf:loss_s}
\end{equation}
\begin{figure*}[htbp]
\includegraphics[width=\linewidth]{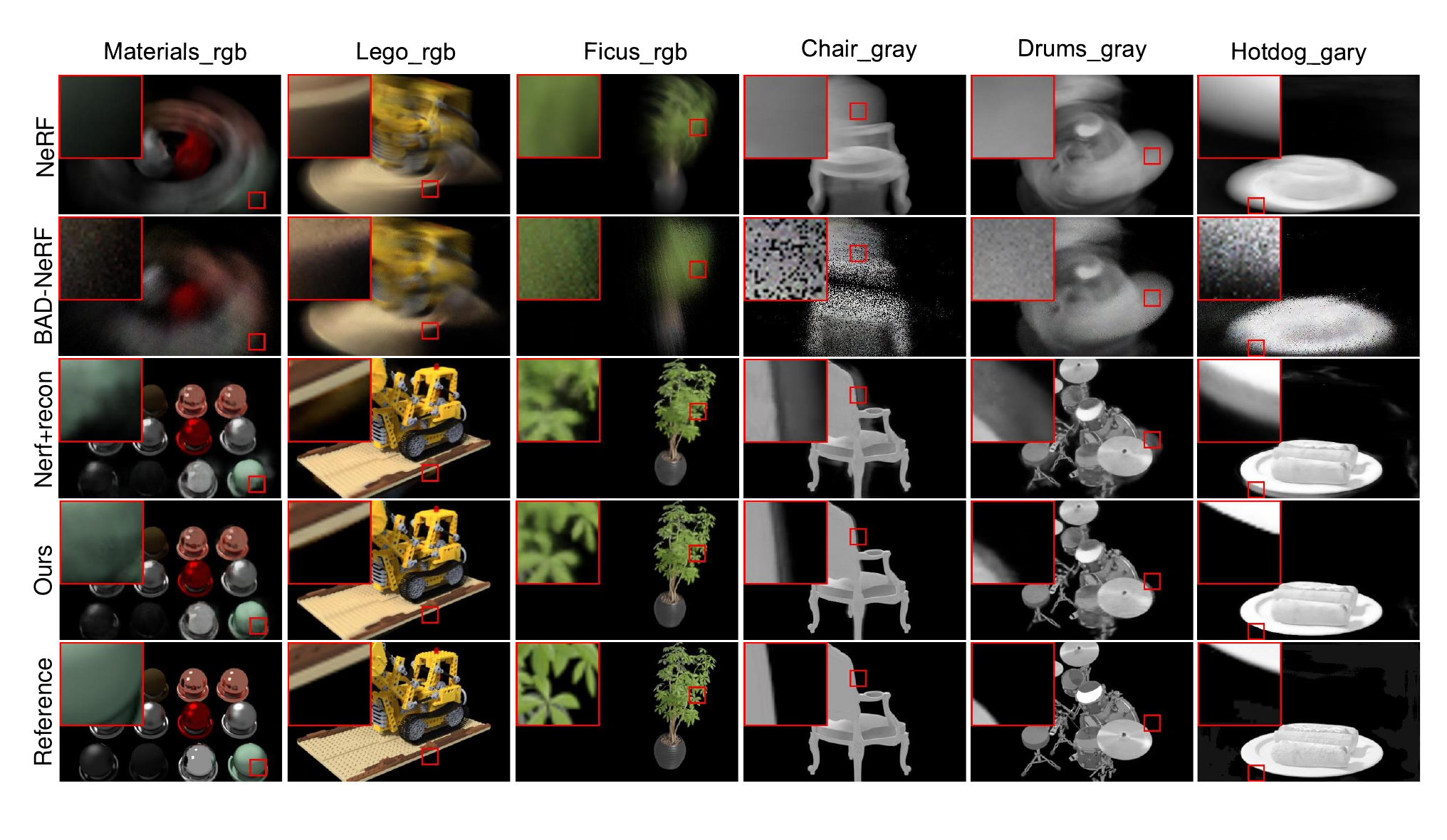}
\centering
\caption{Comparisons on novel view synthesis. We compare our results with three baselines:NeRF, BAD-NeRF, and NeRF+Spk2ImgNet.More details are shown in the green box. NeRF and BAD-NeRF's results have significant blur while NeRF+Spk2ImgNet's results show artifacts. Our results are sharp. The supplement shows more details.}\label{nerf_main_rgb}
\end{figure*}
\subsection{Spike Masks}
\begin{figure}[htb]
    \centering
    \includegraphics[width=\linewidth]{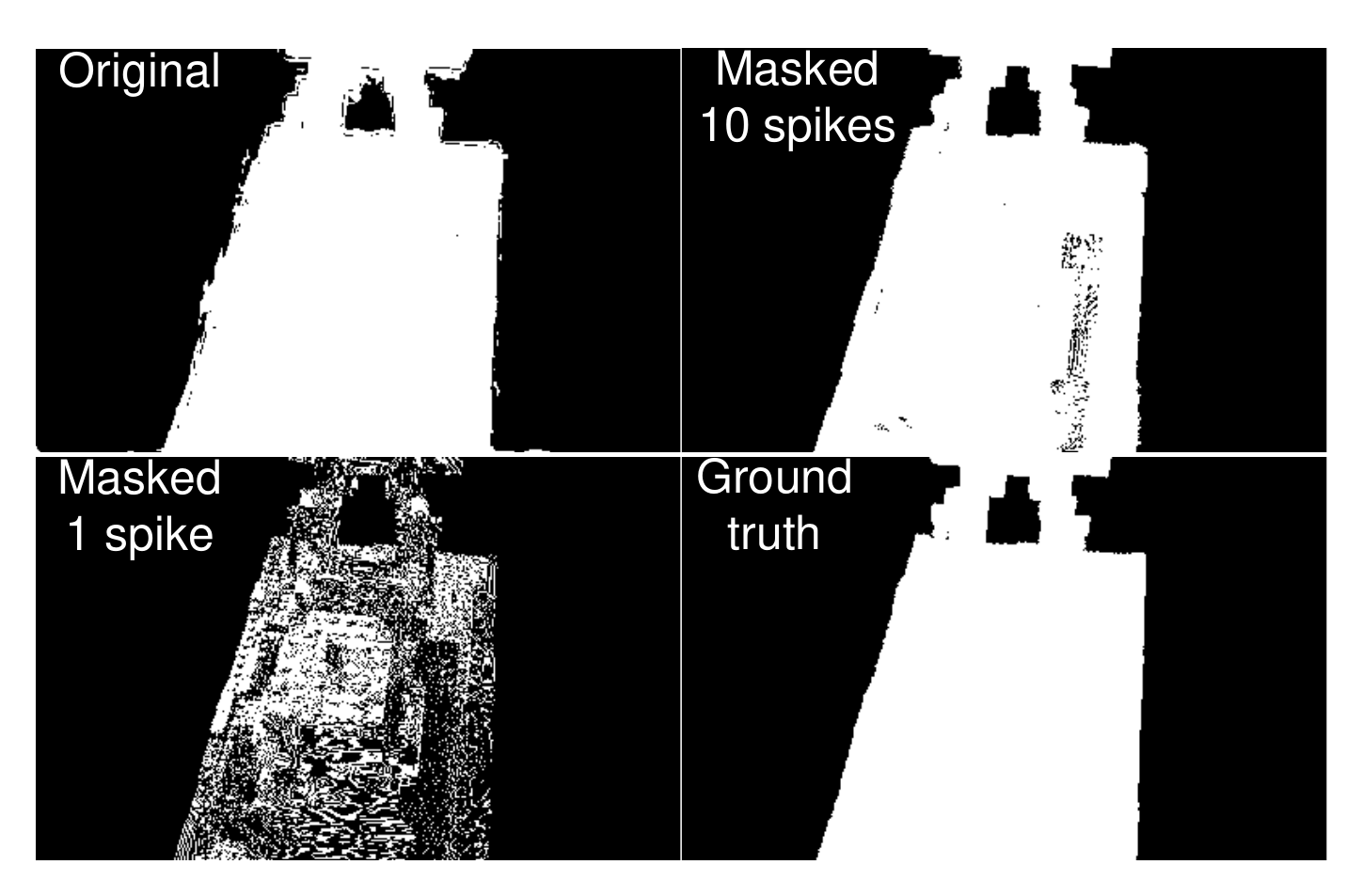}
    \caption{Effective areas (white when r+g+b$>$0 and black when r+g+b=0) for different solutions. Compared with GT, there are obvious error areas when not using masks, and cavities when using single-frame masks. Our solution solves the above problems. }
    \label{mask}
\end{figure}

Due to the serious lack of information in a single spike, all reconstruction methods use multi-frame spikes as input. These methods can reconstruct images with detailed textures from spike streams, but can also introduce erroneous information due to the use of preceding and following frame spikes (see Figure \ref{mask} original), which results in foggy edges in the scene learned by NeRF. We introduce spike masks ${M_s}$ to solve this problem. 
\begin{figure}[htb]
    \centering
    \includegraphics[width=\linewidth]{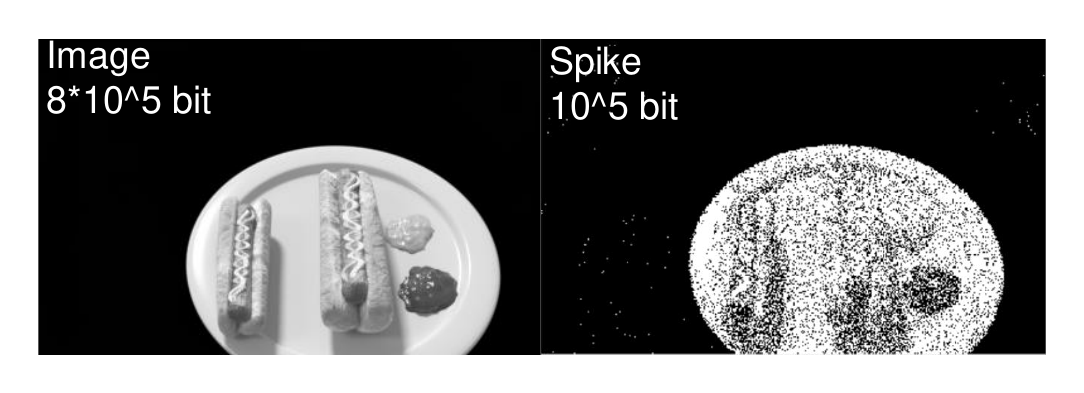}
    \caption{Comparison between spike and image. Compared with images, spikes lack texture details and have a lot of noise because a single frame of spike has less information than an image.}
    \label{bit}
\end{figure}

\begin{table*}[htb]
  \centering

  \caption{ Comparing our method against NeRF, BAD-NeRF and Spk2ImgNet+NeRF on muti scenes. Our
method consistently produces better results. \textbf{Best} results are marked in red and the second best results are marked in yellow.}
    \resizebox{2\columnwidth}{!}{
    \fontsize{22}{28}\selectfont
    \begin{tabular}{ccccccccccccccc}
     \hline
    Dataset& \multicolumn{2}{c}{chair\_rgb} & \multicolumn{2}{c}{drums\_rgb}  & \multicolumn{2}{c}{ficus\_rgb} & \multicolumn{2}{c}{hotdog\_rgb} & \multicolumn{2}{c}{lego\_rgb} & \multicolumn{2}{c}{materials\_rgb} & \multicolumn{2}{c}{average\_rgb}\\
     Method $\vert$ Metrics  & SSIM $\uparrow$& PSNR $\uparrow$ 
                            & SSIM $\uparrow$& PSNR $\uparrow$ 
                            & SSIM $\uparrow$& PSNR $\uparrow$
                            & SSIM$\uparrow$ & PSNR $\uparrow$
                            & SSIM $\uparrow$& PSNR $\uparrow$
                            & SSIM $\uparrow$& PSNR $\uparrow$
                            & SSIM $\uparrow$& PSNR $\uparrow$\\
    
     \cmidrule(r){1-1}\cmidrule(lr){2-3}\cmidrule(lr){4-5}\cmidrule(lr){6-7}\cmidrule(lr){8-9}\cmidrule(lr){10-11}\cmidrule(lr){12-13}\cmidrule(lr){14-15}
     NeRF                   & 0.754 & 21.14 
                            & 0.593 & 21.40
                            & 0.707 & 22.31
                            & 0.752 & 19.31
                            & 0.455 & 17.24
                            & 0.533 & 18.36
                            & 0.632 & 19.96\\
     BAD-Nerf[cvpr23]      & 0.604 & 19.44 
                            & 0.563 & 20.70
                            & 0.597 & 20.97 
                            & 0.658 & 19.42 
                            & 0.385 & 16.02 
                            & 0.397 & 17.56  
                            & 0.534 & 19.02 \\     

     NeRF+Spk2ImgNet[cvpr21]& \cellcolor {yellow!40} 0.961 & \cellcolor {yellow!40} 32.21 
                            & \cellcolor {yellow!40} 0.899 & \cellcolor {yellow!40} 29.90 
                            & \cellcolor {yellow!40} 0.908 & \cellcolor {yellow!40} 27.80 
                            & \cellcolor {yellow!40} 0.920 & \cellcolor {yellow!40} 28.92
                            & \cellcolor {yellow!40} 0.837 & \cellcolor {yellow!40} 26.00
                            & \cellcolor {yellow!40} 0.877 & \cellcolor {yellow!40} 28.27 
                            & \cellcolor {yellow!40} 0.901 & \cellcolor {yellow!40} 28.85\\
    
     Ours              & \cellcolor {red!40} \textbf{0.973} & \cellcolor {red!40} \textbf{32.90}
                            & \cellcolor {red!40} \textbf{0.922} & \cellcolor {red!40} \textbf{30.16}
                            & \cellcolor {red!40} \textbf{0.936} & \cellcolor {red!40} \textbf{29.10} 
                            & \cellcolor {red!40} \textbf{0.923} & \cellcolor {red!40} \textbf{29.69}
                            & \cellcolor {red!40} \textbf{0.861} & \cellcolor {red!40} \textbf{26.32}
                            & \cellcolor {red!40} \textbf{0.912} & \cellcolor {red!40} \textbf{28.67} 
                            & \cellcolor {red!40} \textbf{0.921} & \cellcolor {red!40} \textbf{29.48} \\
    \hline
    Dataset& \multicolumn{2}{c}{chair\_gray} & \multicolumn{2}{c}{drums\_gray}  & \multicolumn{2}{c}{ficus\_gray} & \multicolumn{2}{c}{hotdog\_gray} & \multicolumn{2}{c}{lego\_gray} & \multicolumn{2}{c}{materials\_gray} & \multicolumn{2}{c}{average\_gray}\\
    Methods $\vert$ Metrics  & SSIM $\uparrow$ & PSNR $\uparrow$ 
                            & SSIM $\uparrow$ & PSNR $\uparrow$ 
                            & SSIM $\uparrow$ & PSNR $\uparrow$ 
                            & SSIM$\uparrow$  & PSNR $\uparrow$ 
                            & SSIM $\uparrow$ & PSNR $\uparrow$ 
                            & SSIM $\uparrow$ & PSNR $\uparrow$
                            & SSIM $\uparrow$ & PSNR $\uparrow$ \\
          \cmidrule(r){1-1}\cmidrule(lr){2-3}\cmidrule(lr){4-5}\cmidrule(lr){6-7}\cmidrule(lr){8-9}\cmidrule(lr){10-11}\cmidrule(lr){12-13}\cmidrule(lr){14-15}
    NeRF             & 0.662 & 17.09 
                            & 0.437 & 15.85 
                            & 0.628 & 16.59 
                            & 0.243 & 18.06 
                            & 0.292 & 14.06 
                            & 0.365 & 14.34 
                            & 0.438 & 16.00 \\   
    BAD-Nerf[cvpr23]      & 0.646    & 15.12 
                            & 0.510   & 14.38 
                            & 0.624   & 16.18 
                            & 0.431   & 15.87   
                            & 0.372   & 11.91  
                            & 0.341   & 12.60  
                            & 0.487   & 14.34  \\
     NeRF+Spk2ImgNet[cvpr21]  & \cellcolor {yellow!40} 0.803 & \cellcolor {yellow!40} 27.36  
                            & \cellcolor {yellow!40} 0.671 & \cellcolor {yellow!40} 23.58  
                            & \cellcolor {yellow!40} 0.827 & \cellcolor {yellow!40} 24.79  
                            & \cellcolor {yellow!40} 0.528 & \cellcolor {yellow!40} 25.47  
                            & \cellcolor {yellow!40} 0.636 & \cellcolor {yellow!40} 22.97  
                            & \cellcolor {yellow!40} 0.615 & \cellcolor {yellow!40} 23.12  
                            & \cellcolor {yellow!40} 0.680 & \cellcolor {yellow!40} 24.55   \\

     Ours             & \cellcolor {red!40} \textbf{0.881} & \cellcolor {red!40} \textbf{31.70}  
                            & \cellcolor {red!40} \textbf{0.764} & \cellcolor {red!40} \textbf{25.81}  
                            & \cellcolor {red!40} \textbf{0.874} & \cellcolor {red!40} \textbf{26.18}   
                            & \cellcolor {red!40} \textbf{0.581} & \cellcolor {red!40} \textbf{26.79}  
                            & \cellcolor {red!40} \textbf{0.710} & \cellcolor {red!40} \textbf{25.07}  
                            & \cellcolor {red!40} \textbf{0.769} & \cellcolor {red!40} \textbf{25.43}  
                            & \cellcolor {red!40} \textbf{0.763} & \cellcolor {red!40} \textbf{26.83}  \\
     \hline
    \end{tabular}%
   }
  \label{tab_nerf_main_rgb}%
\end{table*}%

Because of the spatial sparsity of the spike streams, using a single spike mask will lead to a large number of cavities. To address this, we use a relatively small number of multi-frame spikes to fill the cavities. Considering spike $s_i$ at time $t_i$ and reconstruction result $im_i$, we first choose $\mathbb{S}_{ti} = \{ {s}_{j} \in \mathbb{R}^{W \times H} | j = i-n,  i-n+1,\dots\ ,i, \dots\ ,i+n-1,i+n\}$ as original mask sequence. Finally, we have:
\begin{equation}
		 {M_s}=s_{i-n}| s_{i-n+1}| \dots| s_{i+n-1}| s_{i+n}
\end{equation}
where $|$ means or.
After masking image sequence $\mathbb{I}$, we can compute:
\begin{equation}
		L_{recon}= \sum_{r \in R} \Vert M_s(\mathbb{I}(\textbf{x}))-\mathbb{I}_{gt}(\textbf{x}) \Vert\label{nerf_i}
\end{equation}

\section{Experiment}
\label{sec:experiment}
We adopt Novel View Synthesis (NVS) as the standard task to verify our method. We first compare our method with NeRF approaches on traditional cameras and the proposed baseline. We then conduct comprehensive quantitative ablation studies to illustrate the usage of the designed modules. 
\subsection{Implementation Details}
Our code is based on NeRF \cite{nerf} and we train the models for $2*10^5$ iterations on one NVIDIA A100 GPU with the same optimizer and hyper-parameters as NeRF. Since the spiking volume renderer requires continuous multi-spikes, we select the camera pose and sampling points for spiking volume rendering determinedly rather than randomly as NeRF did. We examined our method on synthetic sequences from NeRF \cite{nerf}. We examined six scenes(lego, ficus, chair, materials, hotdog and drums) which cover different conditions. We rendered all of them with a 0.025-second-long 360-degree rotation of the camera around the object resulting in 1000 views to simulate the 40000 fps spike camera and other blurred images in 1000 views to simulate the 400 fps high-speed traditional camera by Blender. Like NeRF, we directly use the corresponding camera intrinsics and extrinnsics generated by the blender. 

\begin{figure}[htb]
\includegraphics[width=\linewidth]
{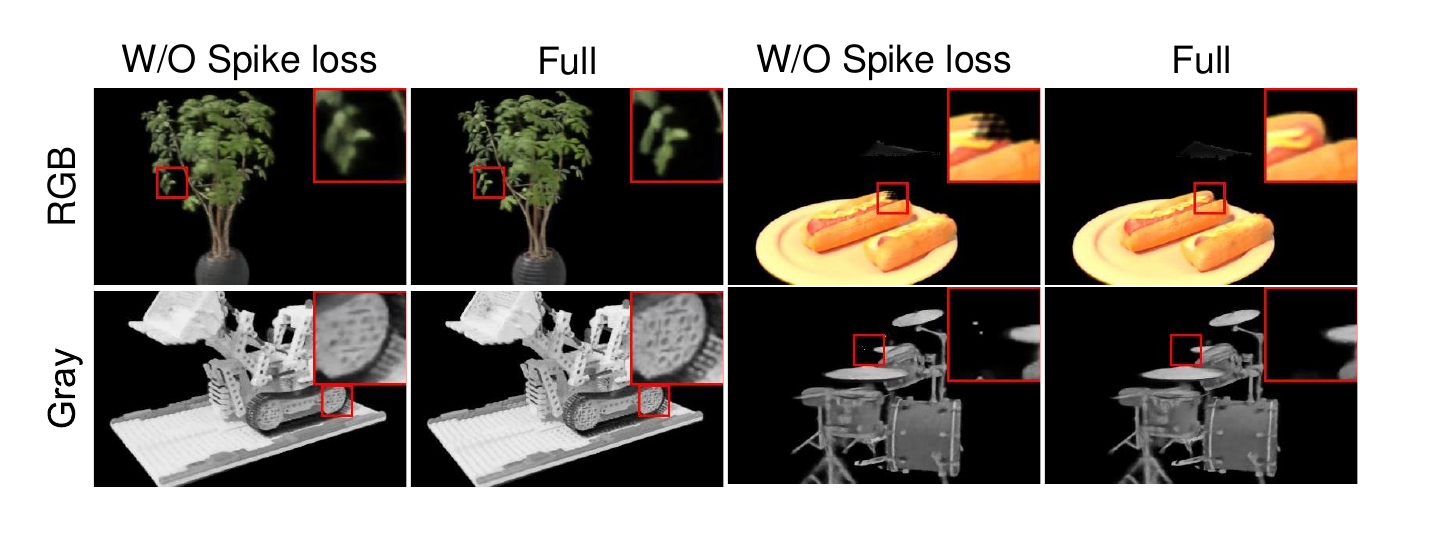}
\centering
\caption{Importance of spike loss. Cavities and blur appear in both RGB and Gray space when disabling spike loss}\label{mask_ab}
\end{figure}
\begin{figure}[htb]
\includegraphics[width=\linewidth]{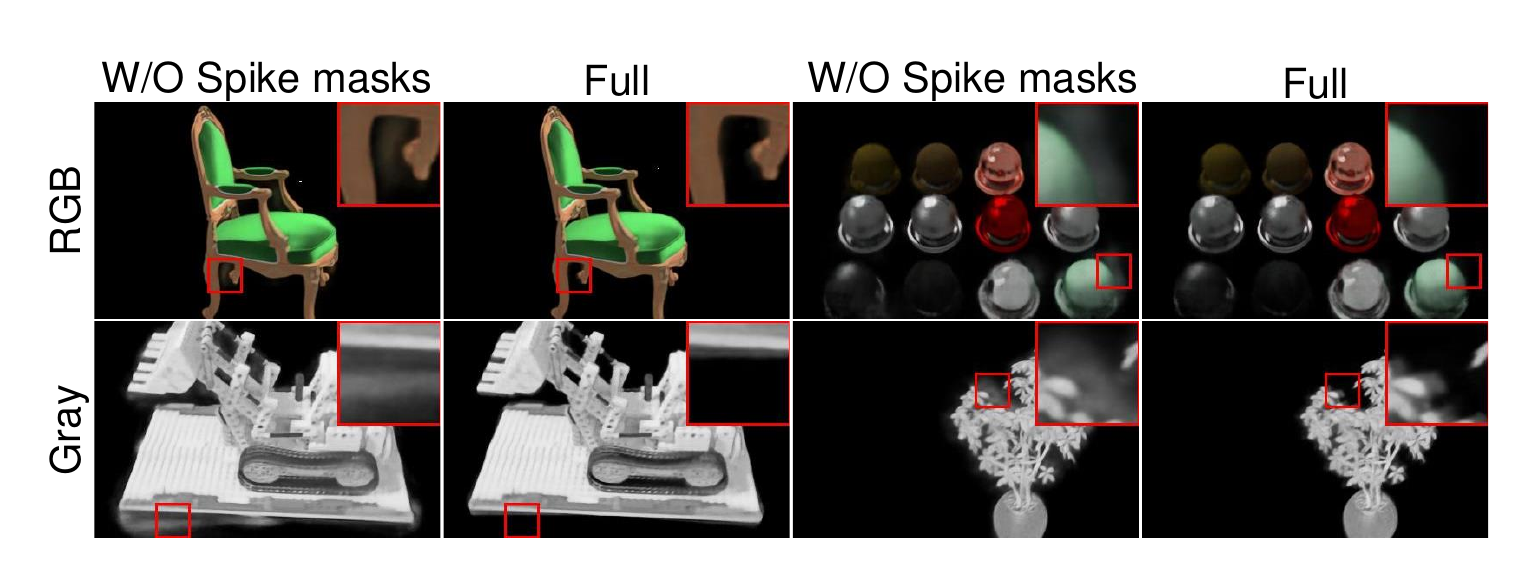}
\centering
\caption{Importance of spike masks. Artifacts appear in both RGB and Gray space when disabling spike masks}\label{spike_ab}
\end{figure}
\subsection{ Comparisons against other Methods}
We compare the spike NeRF results with three baselines: Spk2ImgNet+NeRF\cite{zhao2021spk2imgnet}, the 400 fps traditional camera results on NeRF and BAD-NeRF\cite{wang2023bad}(see Figure \ref{nerf_main_rgb}).To better demonstrate the adaptability of our method to spike cameras, we show both gray and RGB results. From Figure \ref{nerf_main_rgb} we know that our method has absolute advantages over NeRF and BAD-NeRF in high-speed scenes. Compared with directly using spike reconstruction results for training, our method also has obvious supervisors. Corresponding numerical results are reported in Table \ref{tab_nerf_main_rgb} from what we can conclude our method improves more in gray space. 
We also compared the data sizes of two modalities: spikes and images. From Figure
 \ref{bit} we can conclude that spikes have less information resulting in noise and loss of detail. However, our method leverages temporal consistency (see section \ref{sec:method}) to derive stable 3D representations from information-lacking and unstable spike streams.


\begin{table}[htb]
  \centering
  \caption{ Ablation on spike mask and spike loss. \textbf{Best} results are marked in red and the second best results are marked in yellow.}
    \resizebox{1\columnwidth}{!}{
    \fontsize{3.5}{5}\selectfont
    \begin{tabular}{c|ccc}
     \Xhline{0.2px}
   Methods$\vert$Metrics &  SSIM $\uparrow$&  PSNR $\uparrow$& LPIPS  $\downarrow$\\
     \Xhline{0.2px}
     W/O spike masks & 0.899 & 28.97 &\cellcolor {yellow!40} 0.064 \\
     W/O spike loss  & \cellcolor {yellow!40}  0.918 &  \cellcolor {yellow!40} 29.18 & 0.067 \\
    Full  & \cellcolor {red!40} \textbf{0.921} & \cellcolor {red!40} \textbf{29.48} & \cellcolor {red!40} \textbf{0.061} \\
    \Xhline{0.2px}
    \end{tabular}%
   }
  \label{tab_nerf_ab}%
\end{table}%

\subsection{Ablation}

Compared with baselines, our Spike-NeRF introduces two main components: spike masks and the spike volume renderer with spike loss. Next, we discuss their impact on the results. 

\textbf{Spike Loss}:
In section \ref{sec:method}, we proposed spike loss to solve the cavities caused by the partial information loss due to spike masks and the blur caused by incorrect reconstruction. Figure \ref{mask_ab} shows the results before and after disabling spike loss. From Figure \ref{mask_ab} we know that after disabling spike loss, some scenes have obvious degradation in details and a large number of wrong holes. Tab\ref{tab_nerf_ab} shows the improvement of the spike loss.

\textbf{Spike masks}:
Incorrect reconstruction will also lead to a large number of artifacts in NeRF results. We use spike masks to eliminate artifacts to the maximum extent (see \ref{sec:method}).  Figure \ref{spike_ab} shows the results before and after disabling spike masks. From Figure \ref{spike_ab} we know that after disabling spike masks, all scenes have obvious artifacts. Tab\ref{tab_nerf_ab} shows the improvement of the spike masks.

\section{Conclusion}
\label{sec:conclusion}

We introduced the first approach to reconstruct a 3D scene from spike streams which enables photorealistic novel view synthesis in both gray and RGB space. Thanks to the high temporal resolution and unique coding method of the spike camera, Spike-Nerf shows credible advantages in high-speed scenes. Further, we proposed the spiking volume renderer and spike mask so that Spike NeRF outperforms baselines in terms of scene stability and texture details. Our method can also directly generate spike streams. To the best of our knowledge, our paper is also the first time that spike cameras have been used in the field of 3D representation.

\textbf{Limitation:} Due to the difficulty of collecting real spike data with camera poses, Spike-NeRF is only tested on synthetic datasets. In addition, Spike-NeRF assumes that the only moving object in the scene is the spike camera. We believe that NeRF based on spike cameras has greater potential in handling high-speed rigid and non-rigid motions for other objects. Future works can investigate it.
\bibliography{IEEEabrv,main}
\bibliographystyle{IEEEtran}
\newpage
\end{document}